\pdfoutput=1

\documentclass[11pt]{article}

\usepackage{acl}

\usepackage{times}
\usepackage{latexsym}

\usepackage[T1]{fontenc}

\usepackage[utf8]{inputenc}

\usepackage{microtype}

\usepackage[utf8]{inputenc} 
\usepackage[T1]{fontenc}    
\usepackage{hyperref}       
\usepackage{url}            
\usepackage{booktabs}       
\usepackage{amsfonts}       
\usepackage{nicefrac}       
\usepackage{microtype}      
\usepackage{xcolor}         
\usepackage{multirow}
\usepackage{graphicx}
\usepackage{subcaption}
\usepackage[algo2e]{algorithm2e} 

\usepackage{algorithm}  
\usepackage{algpseudocode} 

\usepackage{comment}

\title{AFPQ: Asymmetric Floating Point Quantization for LLMs
}

\author{
Yijia Zhang\textsuperscript{†}\thanks{{\ \ }Equally contributed.}, Sicheng Zhang\textsuperscript{†}\footnotemark[1], Shijie Cao\textsuperscript{‡}, Dayou Du\textsuperscript{§},  
Jianyu Wei\textsuperscript{¶}, Ting Cao\textsuperscript{‡}, Ningyi Xu\textsuperscript{†} \\
\textsuperscript{†}Shanghai Jiao Tong University
\textsuperscript{‡}Microsoft Research Asia \\
\textsuperscript{§}The Hong Kong University of Science and Technology (Guangzhou) \\
\textsuperscript{¶}University of Science and Technology of China \\
\{zhangyijia, zhangsicheng, xuningyi\}@sjtu.edu.cn, \{shijiecao, ting.cao\}@microsoft.com, \\
ddu487@connect.hkust-gz.edu.cn, noob@mail.ustc.edu.cn
}
\begin{document}
\maketitle
\begin{abstract}
Large language models (LLMs) show great performance in various tasks, but face deployment challenges from limited memory capacity and bandwidth.
Low-bit weight quantization can save memory and accelerate inference.
Although floating-point (FP) formats show good performance in LLM quantization, they tend to perform poorly with small group sizes or sub-4 bits.
We find the reason is that the absence of asymmetry in previous FP quantization makes it unsuitable for handling asymmetric value distribution of LLM weight tensors.
In this work, we propose asymmetric FP quantization (AFPQ), which sets separate scales for positive and negative values.
Our method leads to large accuracy improvements and can be easily plugged into other quantization methods, including GPTQ and AWQ, for better performance.
Besides, no additional storage is needed compared with asymmetric integer (INT) quantization.
The code is available at \url{https://github.com/zhangsichengsjtu/AFPQ}.
\end{abstract}

\section{Introduction}

LLMs have significantly advanced language understanding, generation, and reasoning~\cite{touvron2023llama,roziere2023code,zhang2022opt}.
However, the increasing size of LLMs poses great pressure on memory capacity and bandwidth during deployment.
Low-bit quantization is a widely used solution to decrease both memory capacity and bandwidth requirements.
To effectively accommodate LLMs, new quantization methods have been proposed, such as GPTQ~\cite{frantar2022gptq} and AWQ~\cite{lin2023awq}. 
These methods quantize LLMs with low-bit INT formats.

Recent studies suggest utilizing low-bit FP formats, such as FP4 and NF4, in place of INT can lead to improved quantization accuracy of LLMs~\cite{dettmers2023case, zhang2023integer, wu2023zeroquant}. This improvement is attributed to the non-uniform distribution of low-bit FP formats, which more effectively align with LLM weights, characterized by mostly smaller values and a long tail of larger, significant ones.
Although generally superior, FP formats tend to be worse than INT when quantization with small group sizes or sub-4 bits.

\begin{figure}[t]  
    \centering  
    \begin{subfigure}{0.235\textwidth}  
        \includegraphics[width=\textwidth]{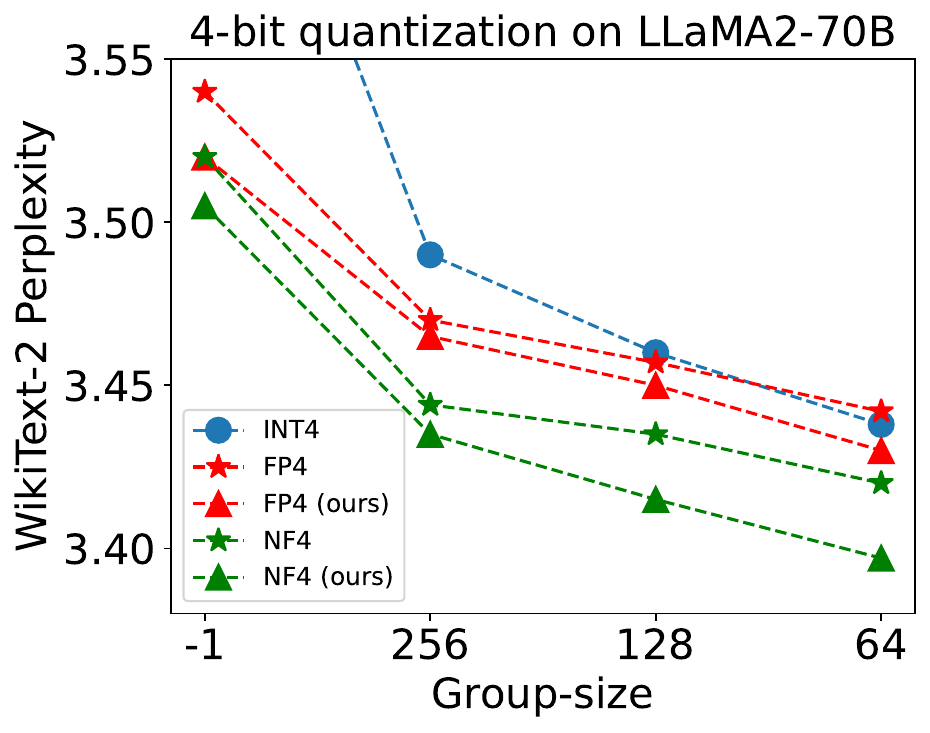}  
    \end{subfigure}  
    \begin{subfigure}{0.23\textwidth}  
        \includegraphics[width=\textwidth]{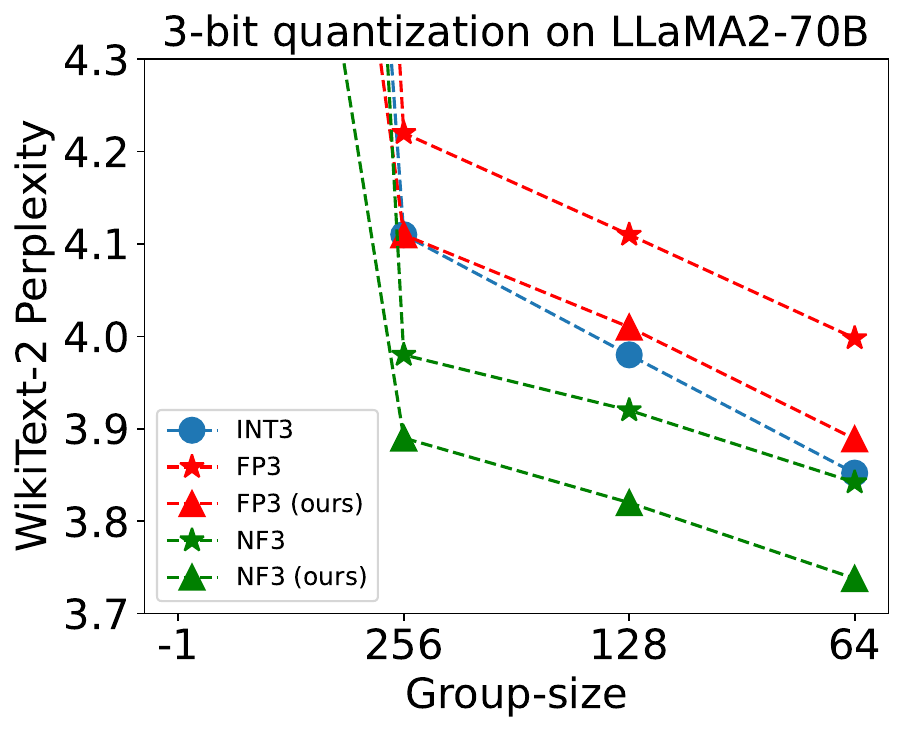}  
    \end{subfigure}  
    \caption{On LLaMA2-70B~\cite{touvron2023llama}, our asymmetric FP quantization reduces the WikiText-2 perplexity (the lower the better) in both 3-bit and 4-bit FP quantization (NF, short for NormalFloat, is an advanced type of FP formats).} 
    \label{fig:fpnfbetter} 
\end{figure}

We identify this is caused by the absence of asymmetry in FP quantization. Given that most weight tensors naturally exhibit asymmetric distributions, it is not suitable to quantize them with standard low-bit FP values, which have a symmetric distribution. Furthermore, we find the conventional methods used in asymmetric INT quantization, such as scale and zero-point adjustments, do not perform well in the context of FP quantization.

In this work, we propose asymmetric floating point quantization (AFPQ), a simple yet effective approach to fit the weight asymmetry in LLMs. 
Unlike previous symmetric FP quantization, which uses a uniform scale for positive and negative values within a weight group, AFPQ sets seperate scales for positive and negative values. AFPQ ensures that the rescaled FP values can better match the original weight values, thereby enhancing quantization accuracy in LLMs.
In Figure~\ref{fig:fpnfbetter}, our AFPQ with FP and NP formats show better results in both 3-bit and 4-bit  round-to-neare (RTN) quantization. Moreover, AFPQ requires no additional storage compared with asymmetric INT quantization. We also validate that the asymmetric FP (FP-asym) low-bit inference system can reach up to 1.62x speedup compared with FP16 systems.

Our contributions can be summarized as follows:

\begin{enumerate}
    \item We identify that the subpar quantization accuracy of FP for LLMs is caused by the asymmetry of weights within the quantization group. 
    \item We introduce the asymmetric FP quantization, which can enhance FP quantization performance significantly.
    \item As AFPQ operates on each individual sub-tensor or group, it can work as a plugin to other tensor-level quantization algorithms, such as GPTQ and AWQ. We integrate asymmetric FP quantization with these methods in this work.
\end{enumerate}

\section{Background and Motivation}

\paragraph{Model quantization methods.} 
Quantization is a process that reduces the precision of Deep Neural Network (DNN) weights to decrease model size and accelerate model inference~\cite{han2015deep,jacob2018quantization}. Existing quantization methods can be broadly categorized into two types: Post Training Quantization (PTQ) and Quantization Aware Training (QAT)~\cite{bengio2013estimating, gholami2022survey}. QAT necessitates model training, which can be expensive, whereas PTQ does not. We focus on PTQ in this work.

\paragraph{Quantization of LLMs.} There are two methods for quantizing LLMs: 1) Quantizing both weights (W) and activations (A), for example, W8A8 quantization~\cite{dettmers2022llm, xiao2023smoothquant}; 2) W-only quantization, for example, W4A16 one~\cite{dettmers2023case}. This article focuses on the W-only method. 
The naive W-only method is RTN.
The advanced methods include GPTQ~\cite{frantar2022gptq} and AWQ~\cite{lin2023awq}. GPTQ uses second-order information to compensate for the error of quantized weights, while AWQ scales salient weights before quantization. Both methods use INT for quantization.

\paragraph{Low-bit Formats.} The current mainstream quantization formats include low-bit INT and FP~\cite{yao2022zeroquant,wu2023zeroquant}. INT is uniformly distributed, while FP, with its exponent and mantissa design, has a distribution that is dense near zero and sparse far from it. In addition, some new formats have also emerged, such as NF~\cite{dettmers20218}, a new type of FP formats designed based on normal number distribution.

\begin{figure}[h]
\centering
\includegraphics[width=0.9\columnwidth]{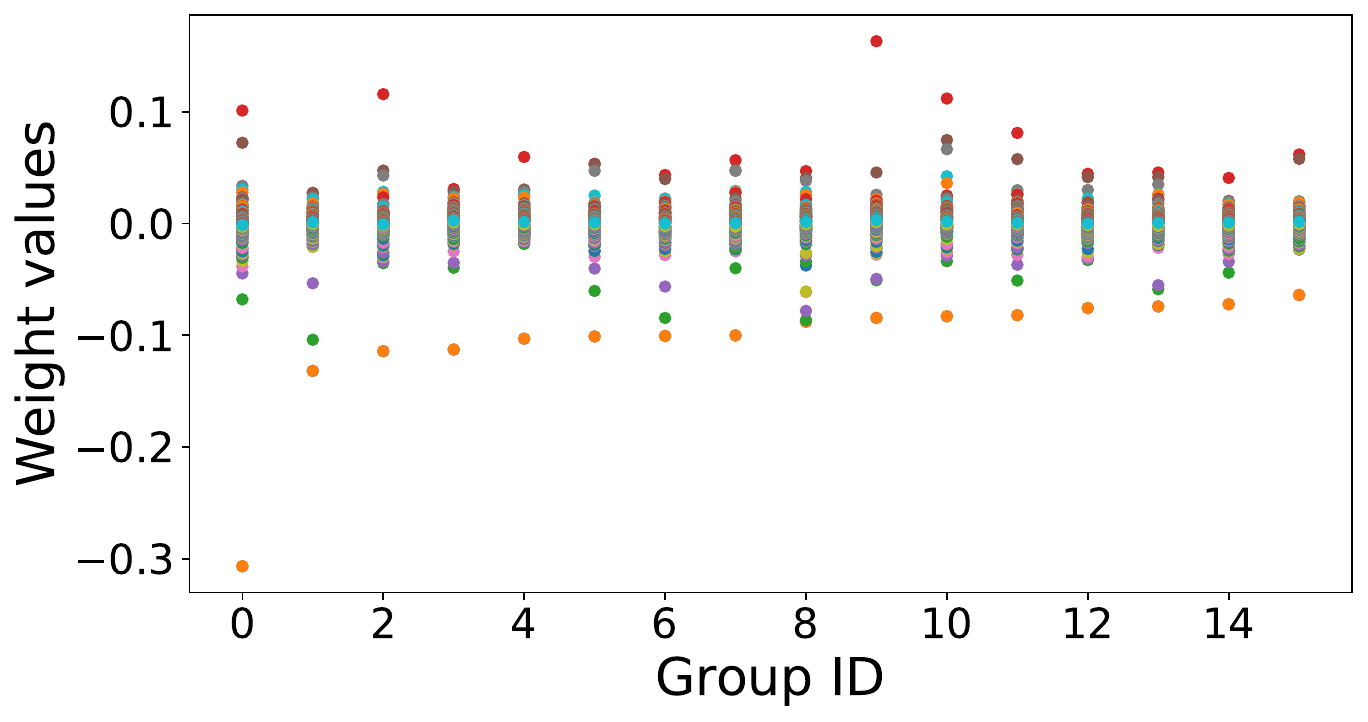}
\caption{Randomly selected weight groups (group-size is 128) from LLaMA2-7B. It is obvious that the maximum and minimum values in many groups are not symmetric about zero.}
\label{fig:max_min}
\end{figure}

\paragraph{Lack of asymmetry for FP quantization}
In the weight tensors of LLMs, outliers often appear~\cite{lin2023awq,dettmers2023spqr}. Due to the randomness of these outliers, many weight tensors exhibit an asymmetric distribution of maximum and minimum values. This phenomenon is particularly noticeable when the group size is small. In Figure~\ref{fig:max_min}, we have randomly selected some LLaMA2 weight groups. It can be observed that more than 50\% of the groups exhibit an asymmetric value distribution.

\begin{figure}[h]
\centering
\includegraphics[width=1\columnwidth]{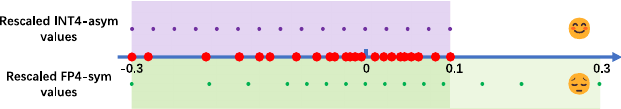}
\caption{Red points are original asymmetric weight values. Recaled INT4-asym covers the weight values well, but the coverage range of rescaled FP4-sym exceeds the range of weight values, thus wasting values in FP formats.}
\label{fig:INTFP}
\end{figure}

For INT, asymmetric quantization with one zero-point (for range translation) and one scale (for scaling) for each weight group can fit the asymmetric tensor distribution well.
For example, if we apply asymmetric INT quantization to asymmetric weights in Figure~\ref{fig:INTFP}, the original weights will be fully covered by the rescaled asymmetric INT (INT-asym) values.
However, when applying previous FP quantization (only one scale for scaling)\footnote{\url{https://github.com/openppl-public/ppq}}\footnote{\url{https://github.com/TimDettmers/bitsandbytes}}, the range of rescaled symmetric FP (FP-sym) values exceeds the range of original weights, leading to a waste of the expressive ability of some FP values. Therefore, asymmetric FP quantization should be introduced for LLMs.

\section{Asymmetric Floating Point Quantization}
\begin{figure}[h]
\centering
\includegraphics[width=1\columnwidth]{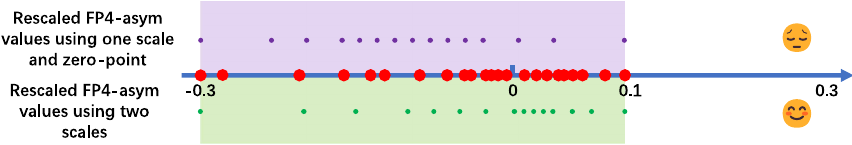}
\caption{Red points are original asymmetric weight values. Recaled FP4-asym using two scales gathers more values near zero than the FP4-asym using one scale and zero-point, which aligns with the distribution of LLMs weights more.}
\label{fig:two_option}
\end{figure}

To make FP quantization applicable to the asymmetric distribution of LLM weights, an intuitive approach is to apply the method with one scale and zero-point used in asymmetric INT quantization to FP quantization, as shown in the purple section of Figure \ref{fig:two_option}. 
However, this approach would shift the dense number area of FP from zero to the left of zero, eliminating the advantages of using FP formats. 
This might make FP less suitable for the value distribution of LLM weights. This phenomenon will be demonstrated in Section 4.

To preserve the advantage of FP formats, we propose asymmetric FP quantization with two separate scales, one for positive numbers and another for negative numbers in each weight group.
In this way, the rescaled FP-asym values from AFPQ can better fit the distribution of original weights, as is shown in Figure~\ref{fig:two_option} green section.
The quantization algorithm is shown in Appendix~\ref{appendix:algorithm}.
The benefits of AFPQ include: 1) Enhanced FP quantization accuracy; 2) No additional storage overhead compared with asymmetric INT quantization (both need two parameters for one group).

As AFPQ operates on each individual sub-tensor or group, it can work as a plugin to other high-level quantization algorithms such GPTQ~\cite{frantar2022gptq} and AWQ~\cite{lin2023awq}.
To demonstrate the applicability, we integrate AFPQ with GPTQ and AWQ for better quantization accuracy for LLMs.
To validate the inference efficiency, we have implemented an low-bit FP-asym inference system.

\section{Experiments} \label{sec:exp}

\textbf{Experimental setup.} We focus on 4-/3-bit PTQ since they can mostly preserve the performance of LLMs~\cite{dettmers2023case}. The formats we use are shown in Appendix~\ref{appendix:formats}. We select LLaMA2~\cite{touvron2023llama} models for basic evaluation because of their superior performance among open-sourced LLMs~\cite{zhang2022opt,scao2022bloom}. We also include WizardCoder~\cite{luo2023wizardcoder} and MetaMath~\cite{yu2023metamath} models for further evaluation. The validation datasets or benchmarks in this section include WikiText-2~\cite{merity2016pointer}, MMLU~\cite{hendryckstest2021}, HumanEval~\cite{chen2021evaluating}, and gsm8k~\cite{cobbe2021gsm8k}. Besides vanilla RTN quantization, we further include experiments based on GPTQ~\cite{frantar2022gptq} and AWQ~\cite{lin2023awq}. We conduct quantization experiments on AutoGPTQ project\footnote{\url{https://github.com/PanQiWei/AutoGPTQ}}.
Our inference system implementation is based on FasterTransformer framework\footnote{\url{https://github.com/NVIDIA/FasterTransformer}}.

\begin{figure}[h]  
    \centering  
    \begin{subfigure}{0.235\textwidth}  
        \includegraphics[width=\textwidth]{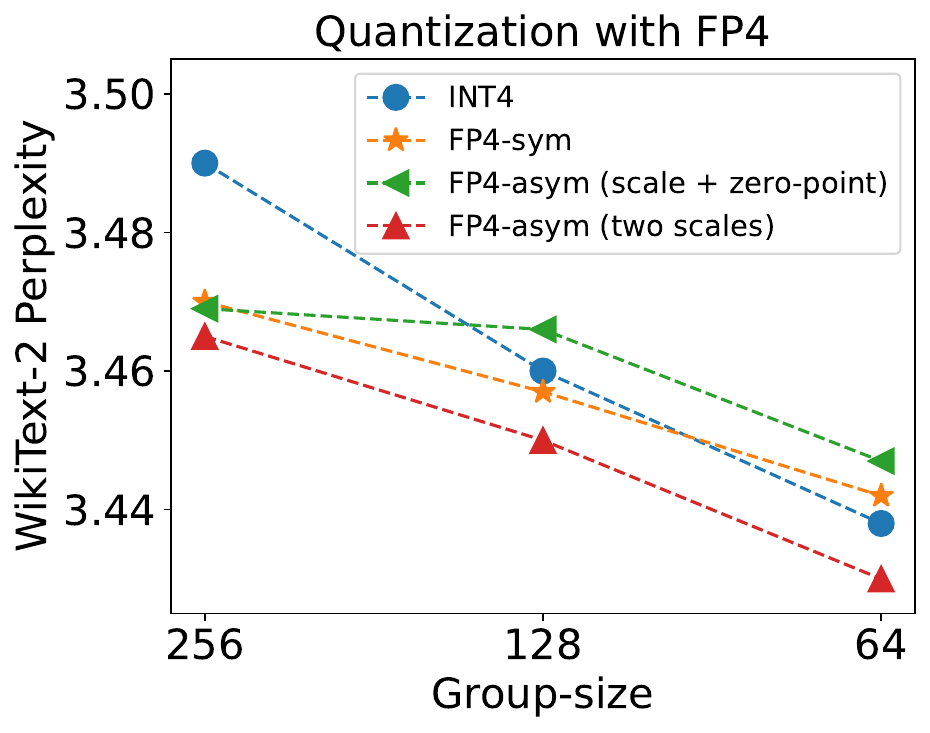}  
    \end{subfigure}  
    \begin{subfigure}{0.23\textwidth}  
        \includegraphics[width=\textwidth]{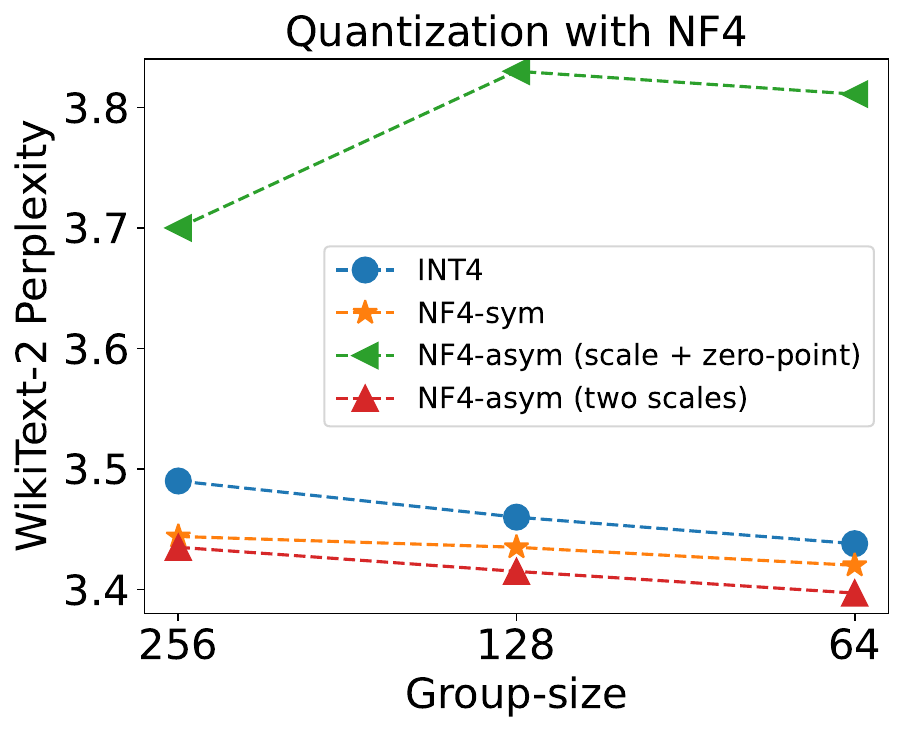}  
    \end{subfigure}  
    \begin{subfigure}{0.23\textwidth}  
        \includegraphics[width=\textwidth]{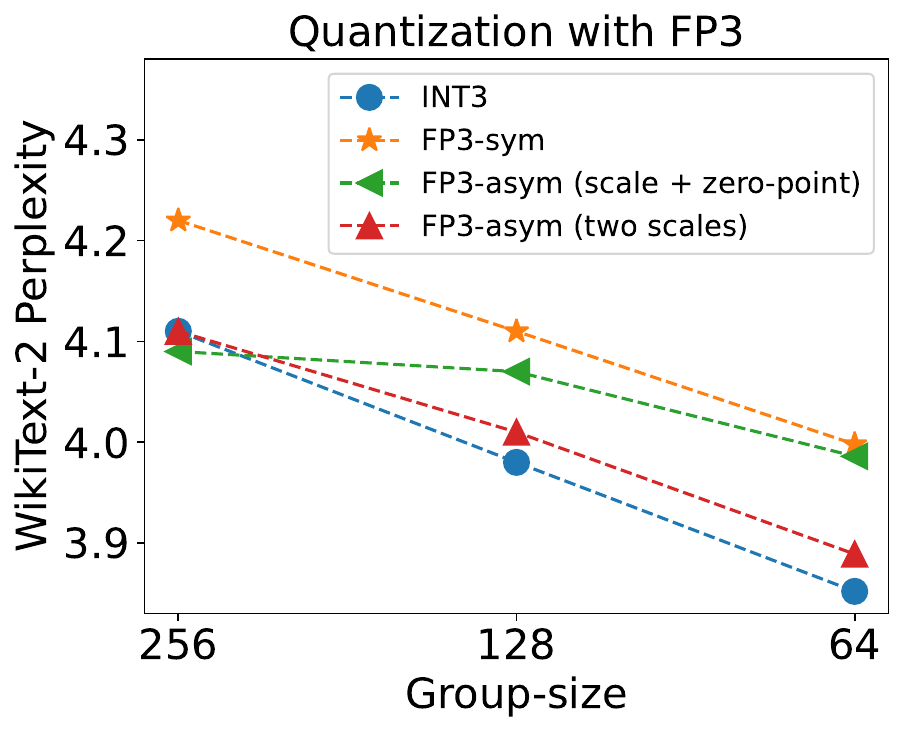}  
    \end{subfigure}
    \begin{subfigure}{0.23\textwidth}  
        \includegraphics[width=\textwidth]{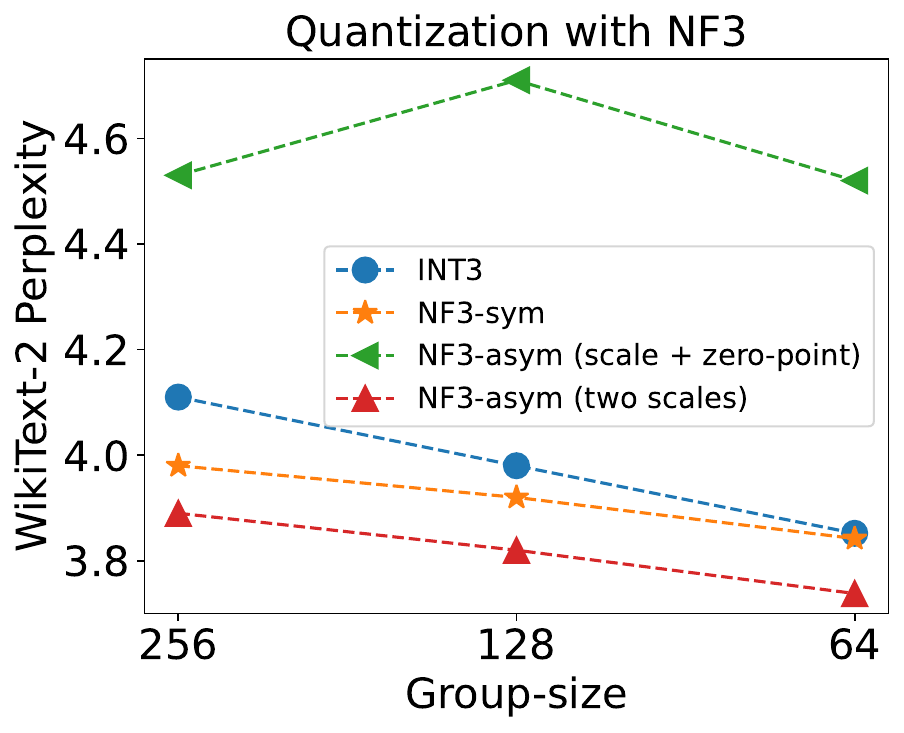}  
    \end{subfigure}
    \caption{When quantizing LLaMA2-70B, FP-asym and NF-asym quantization with two scales shows lower perplexity (ppl) on WikiText-2 (the lower the
    better).} 
    \label{fig:2sbetter} 
\end{figure} 

\begin{table*}[h]
\centering
\caption{WikiText-2 perplexity and MMLU average accuracy on LLaMA2 models after \textbf{4-bit} RTN quantization.}
\resizebox{0.9\textwidth}{!}{%
\begin{tabular}{cc|llll|llll|llll}
\hline
 &  & \multicolumn{4}{c|}{LLaMA2-7B} & \multicolumn{4}{c|}{LLaMA2-13B} & \multicolumn{4}{c}{LLaMA2-70B} \\ \cline{3-14} 
 &  & \multicolumn{1}{c}{g-1} & \multicolumn{1}{c}{g256} & \multicolumn{1}{c}{g128} & \multicolumn{1}{c|}{g64} & \multicolumn{1}{c}{g-1} & \multicolumn{1}{c}{g256} & \multicolumn{1}{c}{g128} & \multicolumn{1}{c|}{g64} & \multicolumn{1}{c}{g-1} & \multicolumn{1}{c}{g256} & \multicolumn{1}{c}{g128} & \multicolumn{1}{c}{g64} \\ \hline
 \multicolumn{1}{c|}{} & FP16 & \multicolumn{4}{c|}{5.47} & \multicolumn{4}{c|}{4.88} & \multicolumn{4}{c}{3.32} \\ \cline{2-14} 
\multicolumn{1}{c|}{\multirow{2}{*}{WikiText-2 $\downarrow$}} & INT4 & 6.12 & 5.75 & 5.72 & 5.67 & 5.20 & 5.02 & 4.98 & 4.97 & 3.67 & 3.49 & 3.46 & 3.44 \\
\multicolumn{1}{c|}{} & NF4-sym & 5.87 & 5.68 & 5.66 & 5.65 & 5.09 & 5.01 & 4.99 & 4.98 & 3.52 & 3.44 & 3.44 & 3.42 \\
\multicolumn{1}{c|}{} & NF4-asym & \textbf{5.77} & \textbf{5.67} & \textbf{5.66} & \textbf{5.64} & \textbf{5.07} & \textbf{5.00} & \textbf{4.98} & \textbf{4.97} & \textbf{3.51} & \textbf{3.44} & \textbf{3.42} & \textbf{3.40} \\ \hline
 \multicolumn{1}{c|}{} & FP16 & \multicolumn{4}{c|}{46.58} & \multicolumn{4}{c|}{55.38} & \multicolumn{4}{c}{69.58} \\ \cline{2-14}
\multicolumn{1}{c|}{\multirow{2}{*}{MMLU(\%) $\uparrow$}} & INT4 & 40.31 & 43.67 & 45.28 & 45.59 & 52.92 & 54.09 & 54.33 & 54.44 & 67.82 & 68.43 & 68.32 & 68.53 \\
\multicolumn{1}{c|}{} & NF4-sym & 43.04 & \textbf{43.94} & 45.09 & 45.70 & 53.59 & 54.37 & 54.58 & 54.84 & \textbf{67.96} & 68.41 & 68.66 & \textbf{69.18} \\
\multicolumn{1}{c|}{} & NF4-asym & \textbf{45.05} & 43.53 & \textbf{45.42} & \textbf{46.12} & \textbf{54.10} & \textbf{54.93} & \textbf{54.71} & \textbf{55.03} & 67.78 & \textbf{68.64} & \textbf{68.81} & 68.93 \\ \hline

\end{tabular}%
}
\label{tab:rtn4bit} 
\end{table*}

\begin{table*}[h]
\centering
\caption{WikiText-2 perplexity and MMLU average accuracy on LLaMA2 models after \textbf{3-bit} RTN quantization. }
\resizebox{0.9\textwidth}{!}{%
\begin{tabular}{cc|cccc|cccc|cccc}
\hline
 &  & \multicolumn{4}{c|}{LLaMA2-7B} & \multicolumn{4}{c|}{LLaMA2-13B} & \multicolumn{4}{c}{LLaMA2-70B} \\ \cline{3-14} 
 &  & g-1 & g256 & g128 & g64 & g-1 & g256 & g128 & g64 & g-1 & g256 & g128 & g64 \\ \hline
\multicolumn{1}{c|}{} & FP16 & \multicolumn{4}{c|}{5.47} & \multicolumn{4}{c|}{4.88} & \multicolumn{4}{c}{3.32} \\ \cline{2-14}
\multicolumn{1}{c|}{\multirow{2}{*}{WikiText-2 $\downarrow$}} & INT3 & 542.80 & 7.10 & 6.66 & 6.40 & 10.68 & 5.67 & 5.52 & 5.39 & 7.53 & 4.11 & 3.98 & 3.85 \\
\multicolumn{1}{c|}{} & NF3-sym & 74.27 & 6.74 & 6.45 & 6.26 & 7.73 & 5.53 & 5.43 & 5.35 & 8.38 & 3.98 & 3.92 & 3.85 \\
\multicolumn{1}{c|}{} & NF3-asym & \textbf{9.85} & \textbf{6.42} & \textbf{6.29} & \textbf{6.15} & \textbf{6.53} & \textbf{5.46} & \textbf{5.35} & \textbf{5.27} & \textbf{5.42} & \textbf{3.89} & \textbf{3.82} & \textbf{3.74} \\ \hline

 \multicolumn{1}{c|}{} & FP16 & \multicolumn{4}{c|}{46.58} & \multicolumn{4}{c|}{55.38} & \multicolumn{4}{c}{69.58} \\ \cline{2-14}
\multicolumn{1}{c|}{\multirow{2}{*}{MMLU(\%) $\uparrow$}} & INT3 & 25.22 & 37.46 & 38.50 & 40.06 & 27.79 & 48.91 & 51.23 & 50.77 & 34.39 & 64.77 & 65.05 & 66.16 \\
\multicolumn{1}{c|}{} & NF3-sym & 26.20 & 36.85 & 38.61 & 38.47 & 38.96 & 49.84 & 50.97 & 51.72 & 40.63 & \textbf{66.40} & 65.90 & \textbf{66.92} \\
\multicolumn{1}{c|}{} & NF3-asym & \textbf{30.31} & \textbf{38.58} & \textbf{41.61} & \textbf{41.11} & \textbf{42.74} & \textbf{52.31} & \textbf{52.60} & \textbf{53.3} & \textbf{56.07} & 66.23 & \textbf{66.78} & 66.43 \\ \hline

\end{tabular}%
}
\label{tab:rtn3bit} 
\end{table*}

\textbf{Comparisons between AFPQ with two scales and the one with scale + zero-point.}
We evaluate LLaMA2-70B with these two methods using the RTN quantization on WikiText-2 perplexity following~\citet{frantar2022gptq}. As shown in Figure~\ref{fig:2sbetter}, quantization using FP-asym with two scales brings better quantization accuracy in both 4-bit and 3-bit grouped quantization for FP and NF formats.
For simplicity, asymmetric FP quantization mentioned below is the one using two scales.
Note that the performance of the FP3 formats is still worse than INT3, this is because FP3 can only represent 7 values for quantization, whereas INT3 and NF3 can represent 8. To ensure a fair comparison, the remaining quantization experiments in this section are conducted using INT and NF formats.

\textbf{Results across various group-sizes and bit-widths using RTN quantization.} To demonstrate the generality of our method, we evaluate our AFPQ using RTN on LLaMA2 models with different bit-widths and group-sizes. The evaluation focuses on WikiText-2 and MMLU benchmark with in-context learning (5-shot) following ~\citet{lin2023awq}. We provide the 4-bit and 3-bit results in Table~\ref{tab:rtn4bit} and Table~\ref{tab:rtn3bit}, respectively. For both bit-widths, quantization with NF-asym achieves better or on-par results in all settings. It performs even better when model size is smaller and bit-width is smaller. For example, NF3-asym with group-size 128 can lead to 3\% MMLU accuracy improvement for LLaMA2-7B (a model size well-suited for edge deployments~\cite{dettmers2023spqr}) compared with INT3 and NF3-sym quantization. The conclusions of FP4 and FP3 are similar to NF formats, which are shown in Appendix~\ref{appendix:FPresults}.

\begin{table}[h]
\centering
\caption{WikiText-2 perplexity and MMLU average accuracy on LLaMA2-70B after we integrate asymmetric FP quantization with \textbf{GPTQ}.}
\resizebox{0.95\columnwidth}{!}{%
\begin{tabular}{cc|llll}
\hline
 &  & \multicolumn{4}{c}{LLaMA2-70B} \\ \cline{3-6} 
 &  & \multicolumn{1}{c}{g-1} & \multicolumn{1}{c}{g256} & \multicolumn{1}{c}{g128} & \multicolumn{1}{c}{g64} \\ \hline
\multicolumn{1}{c|}{\multirow{2}{*}{WikiText-2 $\downarrow$}} &  INT3 & 4.57 & 3.88 & 3.77 & 3.67 \\
\multicolumn{1}{c|}{\multirow{2}{*}{FP16: 3.32}} & NF3-sym & 4.16 & 3.77 & 3.72 & 3.67 \\
\multicolumn{1}{c|}{} & NF3-asym & \textbf{4.07} & \textbf{3.73} & \textbf{3.66} & \textbf{3.61} \\ \hline
\multicolumn{1}{c|}{\multirow{2}{*}{MMLU(\%) $\uparrow$}} & INT3 & 60.10 & 66.65 & 67.25 & 67.75 \\
\multicolumn{1}{c|}{\multirow{2}{*}{FP16: 69.58}} & NF3-sym & 64.45 & 67.03 & 67.42 & 67.72 \\
\multicolumn{1}{c|}{} & NF3-asym & \textbf{64.95} & \textbf{67.33} & \textbf{68.05} & \textbf{68.03} \\ \hline
\end{tabular}%
}
\label{tab:gptq} 
\end{table}

\begin{table}[h]
\centering
\caption{WikiText-2 perplexity and MMLU average accuracy on LLaMA2-70B after we integrate asymmetric FP quantization with \textbf{AWQ}.}
\resizebox{0.95\columnwidth}{!}{%
\begin{tabular}{cc|llll}
\hline
 &  & \multicolumn{4}{c}{LLaMA2-70B} \\ \cline{3-6} 
 &  & \multicolumn{1}{c}{g-1} & \multicolumn{1}{c}{g256} & \multicolumn{1}{c}{g128} & \multicolumn{1}{c}{g64} \\ \hline
\multicolumn{1}{c|}{\multirow{2}{*}{WikiText-2 $\downarrow$}} & INT3 & 4.91 & 4.10 & 3.87 & 3.72 \\
\multicolumn{1}{c|}{\multirow{2}{*}{FP16: 3.32}} & NF3-sym & 4.26 & 4.03 & 3.83 & 3.71 \\
\multicolumn{1}{c|}{} & NF3-asym & \textbf{4.18} & \textbf{3.87} & \textbf{3.74} & \textbf{3.65} \\ \hline
\multicolumn{1}{c|}{\multirow{2}{*}{MMLU(\%) $\uparrow$}} & INT3 & 59.08 & 65.15 & 66.45 & 67.40 \\
\multicolumn{1}{c|}{\multirow{2}{*}{FP16: 69.58}} & NF3-sym & 62.60 & 65.02 & 65.88 & \textbf{67.66} \\
\multicolumn{1}{c|}{} & NF3-asym & \textbf{63.57} & \textbf{66.56} & \textbf{67.00} & 67.41 \\ \hline
\end{tabular}%
}
\label{tab:awq} 
\end{table}

\textbf{Results of applying AFPQ to GPTQ and AWQ.}
Although being effective PTQ methods, there is still an accuracy gap between FP16 LLMs and quantized ones using GPTQ or AWQ. In Table~\ref{tab:gptq} and Table~\ref{tab:awq}, We try to improve these methods by replacing the INT3 quantization with NF3-asym ones in GPTQ and AWQ, respectively. We evaluate LLaMA2-70B with WikiText-2 perplexity and MMLU (5-shot) accuracy. Note that the INT3 or NF3 baseline is already strong, our NF3-asym quantization can still raise the performance to a higher level. 
For group-size 128, the commonly used setting in~\citet{frantar2022gptq,lin2023awq}, our method can reduce WikiText-2 ppl by 0.11 from GPTQ-INT3 and 0.13 from AWQ-INT3, which should be considered significant.

\textbf{Results in coding and mathematical tasks.}
As quantization may hurt LLMs' performance in difficult downstream tasks, such as coding and mathematical ones, we also evaluate AFPQ on the WizardCoder-7B model and the MetaMath-7B model in Table~\ref{tab:mathcode}. The benchmark for WizardCoder and MetaMath is HumanEval and gsm8k, respectively. We use AWQ with NF3-asym in the group-size-64 quantization. We can see that NF3-asym helps reach the highest quantization accuracy in both tasks. Notably, the accuracy of quantized WizardCoder-7B is enhanced by 4.87\% compared with AWQ-INT3, which strongly proves the effectiveness of our method.

\begin{table}[h]
\centering
\caption{Evaluation results on WizardCoder-7B and MetaMath-7B after 3-bit AWQ with group-size of 64. For WizardCoder-7B, we show the percentage of pass rates on the HumanEval. For MetaMath-7B, we show the testing accuracy on gsm8k.}
\resizebox{0.9\columnwidth}{!}{%
\begin{tabular}{c|cccc}
\hline
 & FP16 & INT3 & NF3-sym & NF3-asym \\ \hline
WizardCoder-7B $\uparrow$ & 57.31 & 47.56 & 45.12 & \textbf{52.43} \\
MetaMath-7B $\uparrow$ & 66.41 & 63.52 & 60.86 & \textbf{64.53} \\ \hline
\end{tabular}%
}
\label{tab:mathcode} 
\end{table}

\textbf{Efficiency evaluation.}
Since our AFPQ method needs to store two parameters (two scales) for each quantization group, the same as the asymmetric INT quantization (one scale and one zero-point), no additional storage is needed for our method compared with the INT-asym one. As for the inference speed, since low-bit NF-based kernels have not been proposed in previous work, we develop these kernels and integrate them into FasterTransformer framework. The implementation details can be found in Appendix~\ref{appendix:kernel}. We measure the end-to-end latency of LLaMA2 models on a single A6000 GPU. We keep the batch size to be 1, the input sequence length to be 128, and a uniform output token count of 20. In Table~\ref{tab:speedup}, our AFPQ method with NF4-asym achieves up to 1.62x speedup compared with FP16 baseline. Although it incurs inference overhead compared with INT4-/NF4-sym-based system, we believe the gap can be narrowed with kernel optimizations, which we leave it as a future work.

\begin{table}[h]
\centering
\caption{Inference latency (ms) of LLaMA2-7B and LLaMA2-13B under different formats}
\resizebox{0.9\columnwidth}{!}{%
\begin{tabular}{c|cccc}
\hline
 & FP16 & INT4 & NF4-sym & NF4-asym \\ \hline
LLaMA2-7B & 415.06 & 174.29 & 187.23 & 265.54 \\
LLaMA2-13B & 788.01 & 309.87 & 317.15 & 485.42 \\ \hline
\end{tabular}%
}
\label{tab:speedup} 
\end{table}

\section{Conclusion}
In this study, we identify that the lack of asymmetry in previous FP quantization can lead to poor quantization for LLM weight tensors with asymmetric distribution.
To solve the problem, we propose asymmetric FP quantization which sets separate scales for positive and negative values.
Our method can be easily plugged into other effective methods, including GPTQ and AWQ, for performance improvements.
AFPQ enhances LLM quantization results and needs no additional storage compared with asymmetric INT quantization.

\bibliography{custom}
\bibliographystyle{acl_natbib}

\newpage

\appendix
\section*{Appendix}
\section{Low-bit formats used in this work} \label{appendix:formats}
In this work, we use FP4 E2M1 and FP3 E2M0 formats. Both excludes NaN and Inf following~\citet{zhang2023integer}. For NF formats, we use the values from Bitsandbytes\footnote{\url{https://github.com/TimDettmers/bitsandbytes}}.
The exact values of the INT, FP and NF formats used in our experiments are as follows: 

\textbf{INT4:} [-8, -7, -6, -5, -4, -3, -2, -1, 0, 1, 2, 3, 4, 5, 6, 7]

\textbf{FP4:} [-6, -4, -3, -2, -1.5, -1, -0.5, 0, 0.5, 1, 1.5, 2, 3, 4, 6]

\textbf{NF4:} [-1, -0.6961928009986877, -0.5250730514526367, -0.39491748809814453, -0.28444138169288635, -0.18477343022823334, -0.09105003625154495, 0, 0.07958029955625534, 0.16093020141124725, 0.24611230194568634, 0.33791524171829224,  0.44070982933044434, 0.5626170039176941, 0.7229568362236023, 1]

\textbf{INT3:} [-4, -3, -2, -1, 0, 1, 2, 3]

\textbf{FP3:} [-4, -2, -1, 0, 1, 2, 4]

\textbf{NF3:} [-1, -0.5350227355957031, -0.2469314038753510, 0, 0.1833375245332718, 0.3819939494132996, 0.6229856610298157, 1]

\section{Quantization algorithm} \label{appendix:algorithm}
We present the pseudocode of the quantization algorithm used in our experiments in Algorithm~\ref{alg:quant_method}, including existing INT-based algorithm and FP-based algorithm.
\begin{algorithm}
  \SetKwFunction{INTQuant}{INTQuant}\SetKwFunction{INTDequant}{INTDequant}
  \SetKwFunction{FPSYMQuant}{FPSYMQuant}\SetKwFunction{FPSYMDequant}{FPSYMDequant}
  \SetKwFunction{FPASYMQuant}{FPASYMQuant}\SetKwFunction{FPASYMDequant}{FPASYMDequant}
  \SetKwInOut{Input}{input}\SetKwInOut{Output}{output}

    // \textit{$weight_{max}$ represents the maximal value in each weight group, $range$ represents the range of quantization formats}
  
    // \textit{INT-based algorithm}
    
    def \INTQuant:

        \For{weight group in weight tensor to quantize} {
        
        $scale = \frac{weight_{max} - weight_{min}}{range}$
        
        $zeropoint = [\frac{-weight_{min}}{scale}]$
        
        $weight\_4bit = [\frac{weight}{scale}] + zeropoint$
        }
        \BlankLine
    def \INTDequant:

        \For{weight group in weight tensor to quantize} {
        $weight\_q = scale (weight\_4bit - zeropoint)$
        }
        \BlankLine
        \hrule
        \BlankLine
        
    // FP-based symmetric algorithm
    
    def \FPSYMQuant:

        \For{weight group in weight tensor to quantize} {
        
        $scale = \frac{max(weight_{max}, |weight_{min}|)}{range/2}$
        
        $weight\_4bit = [\frac{weight}{scale}]$
        }
        \BlankLine
    def \FPSYMDequant:

        \For{weight group in weight tensor to quantize} {
        $weight\_q = scale * weight\_4bit$
        }
        \BlankLine
        \hrule
        \BlankLine

    // FP-based asymmetric algorithm
    
    def \FPASYMQuant:

        \For{weight group in weight tensor to quantize} {
        $scale_{pos} = \frac{weight_{max}}{range/2}$

        $scale_{neg} = \frac{-weight_{min}}{range/2}$
        
        $weight\_4bit = [\frac{weight_{pos}}{scale_{pos}}] + [\frac{weight_{neg}}{scale_{neg}}]$
        }
        \BlankLine
    def \FPASYMDequant:

        \For{weight group in weight tensor to quantize} {
        $weight\_q = scale_{pos} * weight\_4bit_{pos} + scale_{neg} * weight\_4bit_{neg}$
        }

  \caption{Basic quantization methods}
  \label{alg:quant_method}
\end{algorithm}

\section{Results of AFPQ with FP formats} \label{appendix:FPresults}
Additional results of RTN quantization using FP4/3 formats are shown in Table~\ref{tab:rtn4bitfull} and Table~\ref{tab:rtn3bitfull}, respectively.

\begin{table*}[h]
\caption{WikiText-2 perplexity and MMLU average accuracy on LLaMA2 models after FP4 RTN quantization}
\resizebox{\textwidth}{!}{%
\begin{tabular}{cc|llll|llll|llll}
\hline
 &  & \multicolumn{4}{c|}{LLaMA2-7B} & \multicolumn{4}{c|}{LLaMA2-13B} & \multicolumn{4}{c}{LLaMA2-70B} \\ \cline{3-14} 
 &  & \multicolumn{1}{c}{g-1} & \multicolumn{1}{c}{g256} & \multicolumn{1}{c}{g128} & \multicolumn{1}{c|}{g64} & \multicolumn{1}{c}{g-1} & \multicolumn{1}{c}{g256} & \multicolumn{1}{c}{g128} & \multicolumn{1}{c|}{g64} & \multicolumn{1}{c}{g-1} & \multicolumn{1}{c}{g256} & \multicolumn{1}{c}{g128} & \multicolumn{1}{c}{g64} \\ \hline
 \multicolumn{1}{c|}{} & FP16 & \multicolumn{4}{c|}{5.47} & \multicolumn{4}{c|}{4.88} & \multicolumn{4}{c}{3.32} \\ \cline{2-14}

\multicolumn{1}{c|}{\multirow{2}{*}{WikiText-2 $\downarrow$}} & INT4 & 6.12 & 5.75 & 5.72 & 5.67 & 5.20 & 5.02 & 4.98 & 4.97 & 3.67 & 3.49 & 3.46 & 3.44 \\
\multicolumn{1}{c|}{} & FP4-sym & 5.89 & 5.73 & 5.70 & 5.67 & 5.11 & 5.03 & 5.02 & 5.01 & 3.54 & 3.47 & 3.46 & 3.44 \\
\multicolumn{1}{c|}{} & FP4-asym & 5.82 & 5.71 & 5.70 & 5.67 & 5.09 & 5.02 & 5.01 & 4.99 & 3.52 & 3.47 & 3.45 & 3.43 \\ \hline

 \multicolumn{1}{c|}{} & FP16 & \multicolumn{4}{c|}{46.58} & \multicolumn{4}{c|}{55.38} & \multicolumn{4}{c}{69.58} \\ \cline{2-14}
\multicolumn{1}{c|}{\multirow{2}{*}{MMLU(\%) $\uparrow$}} & INT4 & 40.31 & 43.67 & 45.28 & 45.59 & 52.92 & 54.09 & 54.33 & 54.44 & 67.82 & 68.43 & 68.32 & 68.53 \\
\multicolumn{1}{c|}{} & FP4-sym & 44.14 & 44.25 & 43.74 & 44.04 & 53.77 & 54.17 & 54.83 & 54.62 & 68.14 & 68.72 & 68.71 & 68.90 \\
\multicolumn{1}{c|}{} & FP4-asym & 45.25 & 44.61 & 45.15 & 44.55 & 54.23 & 54.47 & 54.70 & 54.99 & 68.74 & 68.65 & 68.86 & 69.06 \\ \hline

\end{tabular}%
}
\label{tab:rtn4bitfull} 
\end{table*}

\begin{table*}[h]
\caption{WikiText-2 perplexity and MMLU average accuracy on LLaMA2 models after FP3 RTN quantization}
\resizebox{\textwidth}{!}{%
\begin{tabular}{cc|cccc|cccc|cccc}
\hline
 &  & \multicolumn{4}{c|}{LLaMA2-7B} & \multicolumn{4}{c|}{LLaMA2-13B} & \multicolumn{4}{c}{LLaMA2-70B} \\ \cline{3-14} 
 &  & g-1 & g256 & g128 & g64 & g-1 & g256 & g128 & g64 & g-1 & g256 & g128 & g64 \\ \hline
 \multicolumn{1}{c|}{} & FP16 & \multicolumn{4}{c|}{5.47} & \multicolumn{4}{c|}{4.88} & \multicolumn{4}{c}{3.32} \\ \cline{2-14}

\multicolumn{1}{c|}{\multirow{2}{*}{WikiText-2 $\downarrow$}} & INT3 & 542.80 & 7.10 & 6.66 & 6.40 & 10.68 & 5.67 & 5.52 & 5.39 & 7.53 & 4.11 & 3.98 & 3.85 \\
\multicolumn{1}{c|}{} & FP3-sym & 1621.90 & 7.16 & 6.89 & 6.64 & 12.76 & 5.82 & 5.66 & 5.54 & 8.43 & 4.22 & 4.11 & 4.00 \\
\multicolumn{1}{c|}{} & FP3-asym & 18.72 & 6.89 & 6.63 & 6.48 & 7.72 & 5.69 & 5.57 & 5.41 & 5.93 & 4.11 & 4.01 & 3.89 \\ \hline

 \multicolumn{1}{c|}{} & FP16 & \multicolumn{4}{c|}{46.58} & \multicolumn{4}{c|}{55.38} & \multicolumn{4}{c}{69.58} \\ \cline{2-14}
\multicolumn{1}{c|}{\multirow{2}{*}{MMLU(\%) $\uparrow$}} & INT3 & 25.22 & 37.46 & 38.50 & 40.06 & 27.79 & 48.91 & 51.23 & 50.77 & 34.39 & 64.77 & 65.05 & 66.16 \\
\multicolumn{1}{c|}{} & FP3-sym & 23.73 & 31.75 & 36.55 & 33.08 & 27.13 & 48.66 & 49.76 & 49.89 & 32.32 & 64.65 & 65.17 & 65.91 \\
\multicolumn{1}{c|}{} & FP3-asym & 27.32 & 35.42 & 40.33 & 40.24 & 36.15 & 50.09 & 50.72 & 51.60 & 49.74 & 64.62 & 66.14 & 66.41 \\ \hline

\end{tabular}%
}
\label{tab:rtn3bitfull} 
\end{table*}

\section{Kernel implementation} \label{appendix:kernel}
Currently, W-only quantization requires low-bit weights to be dequantized to FP16 during inference, and then calculations are performed with the FP16 activations. 
In our system implementation, we store two 4-bit quantized weights using one byte. During dequantization, we load the byte and recover it to two 4-bit weights. 

For INT and FP formats, the conversion from 4-bit to FP16 values can be completed by algebraic computation. For NP formats, it can be realized by using look-up tables (LUTs). 
Then these values can be further dequantized using the methods in Algorithm\ref{alg:quant_method}.

\end{document}